\documentclass{article}
\usepackage{arxiv}

\usepackage[utf8]{inputenc} 
\usepackage[T1]{fontenc}    
\usepackage{hyperref}       
\usepackage{url}            
\usepackage{booktabs}       
\usepackage{amsfonts}       
\usepackage{nicefrac}       
\usepackage{microtype}      
\usepackage{lipsum}
\usepackage{pdflscape}
\usepackage{afterpage}
\usepackage{capt-of}
\usepackage{array}
\usepackage{rotating}
\usepackage{float}
\usepackage{amsmath}

\usepackage[english]{babel}
\usepackage{csquotes}
\usepackage{biblatex}
\usepackage{graphicx}
\graphicspath{ {./images/} }

\title{Intrinsic analysis for dual word embedding space models}

\author{
  Mohit Mayank\\
  Tata Consultancy Service, Pune, India\\
  \texttt{m.mohit@tcs.com} \\
}
\date{}
\begin{document}
\maketitle

\begin{abstract}
Recent word embeddings techniques represent words in a continuous vector space, moving away from the atomic and sparse representations of the past. Each such technique can further create multiple varieties of embeddings based on different settings of hyper-parameters like embedding dimension size, context window size and training method. One additional variety appears when we especially consider the Dual embedding space techniques which generate not one but two-word embeddings as output. This gives rise to an interesting question - "\textit{is there one or a combination of the two word embeddings variety, which works better for a specific task?}". This paper tries to answer this question by considering all of these variations. Herein, we compare two classical embedding methods belonging to two different methodologies - \textit{Word2Vec} from window-based and \textit{Glove} from count-based. For an extensive evaluation after considering all variations, a total of 84 different models were compared against semantic, association and analogy evaluations tasks which are made up of 9 open-source linguistics datasets. The final \textit{Word2vec} reports showcase the preference of non-default model for 2 out of 3 tasks. In case of \textit{Glove}, non-default models outperform in all 3 evaluation tasks.
\end{abstract}


\section{Introduction}
\label{sec:introduction}

Representing word as a continuous vector traces its history back to the last century where one of the most promising works used back-propagation error to learn the vector representations \cite{de1986learning}. Over the years several methods were proposed and they can be categorised into two different approaches. First, the count-based approach - where some variety of global word-context co-occurrence matrix is leveraged to learn the vector representations \cite{rohde2006improved, collobert2014word, pennington2014glove}. Second, the shallow window-based methods - where a sliding window approach is applied on a large corpus of text and embeddings are learned by considering the context words within a predefined local window of the target word \cite{mikolov2013distributed, mikolov2013efficient, levy-goldberg-2014-linguistic}. Both of the generalisations, on the lowest level, learn the embedding by considering multiple pairs of words. While the internal mechanics and the heuristics considered is different, overall the approaches perform the same task of training word embedding such that the linguistic relationship between the pair of words mirrors the spatial relationship between vectors embeddings of the words. The initial push for creating word embeddings was to move away from the atomic and sparse nature of words to a more continuous one, as it leads to smaller size representation of words which saves storage. Another interesting aspect is the formation of latent features which is represented by each dimension of the learned continuous word vectors. This highlights the notion that there are multiple perspectives to look at a word and a pair of words can be similar by any combination of these perspectives. For example, a word embedding could have the first dimension store the notion of \textit{royalty}, second dimension the \textit{gender}, third the \textit{age} and so on. And two words like \textit{king} and \textit{queen} may be very similar based on the first dimension of royalty but completely different from the second dimension of gender. This is because while training word pairs which are usually seen together are forced to have similar vector representation.

Recent works extend this idea to highlight that these embeddings also showcase several semantic and morphological properties. This is a further generalization of latent features learning and proposes unique applications of the embeddings. For examples, word embedding can be used to find \textit{semantic similarity} between words, where a simple vector similarity method like cosine similarity between the embeddings of two words also represents how similar two words are, like in the case where $frog$ and $toad$ are more semantically similar than $frog$ and $cat$. Another type could be \textit{association or relatedness similarity} where $cup$ and $cupboard$ are more related to each other and hence have higher similarity than say $cup$ and $car$. Hence word embedding, irrespective of their origins, can be used for a wide variety of NLP tasks. This led to the requirement for a fast and empirical evaluation of the embeddings, which could suggest its performance for different kind of tasks. Existing evaluation schemes fall into two major categories:
extrinsic and intrinsic evaluation. In extrinsic evaluation, we use word embeddings as input features to a downstream task and measure changes in performance metrics specific to that task. Examples include part-of-speech tagging and named-entity
recognition \cite{pennington2014glove}. Intrinsic evaluations on the other hand directly test for syntactic or semantic relationships between words \cite{mikolov2013distributed, mikolov2013efficient}. These tasks typically involve a pre-selected set of query terms and
semantically related target words. Methods are evaluated by compiling an aggregate score for each method such as a correlation coefficient, which then serves as an absolute measure of quality. While the extrinsic evaluation provides a reality check and connects the real world downstream application, it's quite expensive as additional training specific to the task is required to be performed other than the embedding training. On the other hand, intrinsic evaluation is fast, as it only needs embedding training, but falls shy by providing just a simulation of the real downstream application. That said, intrinsic evaluation has been the baseline for embedding research and provides a holistic understanding of the embeddings. Due to this reason, we prefer intrinsic evaluation by considering several datasets covering wide variety of intrinsic evaluations such as \textit{similarity  evaluation}, \textit{association evaluation} and \textit{analogy evaluation}.

There has been extensive research on modifying the existing techniques by proposing sophisticated newer models which perform relatively better on a particular evaluation task. Despite the growing interest in creating newer complex vector representations models, there has been relatively little work on direct evaluations of different internal varieties of the existing embedding models. One such variety is exposed when considering the fact that classical models like Glove \cite{pennington2014glove} and Word2Vec  \cite{mikolov2013distributed} train not one but two word embeddings, but returns only one as final embeddings. This final word embedding could be the sum of both (in Glove) or the first one (in Word2vec). The availability of dual embedding vectors and that a selection operation is involved raises several valid questions on different selection operations for the final embeddings. The exploration to find the best internal variety of existing model is further complicated when considering the vast possibilities of hyper-parameters values of the models \cite{levy2015improving}. Without proper investigation, this leaves the task of best model selection mostly as a random guess rather than being an empirical understanding. Also finding the best model currently requires extensive evaluation which is computationally expensive. 
 
This paper explores this problem in detail and the major contributions can be listed as follows, 
\begin{enumerate}
\item The paper highlights the problem of ignoring internal varieties of existing embedding models. 
\item The paper performs an extensive analysis to propose suitable internal variety for different tasks for different models.
\item The paper considers the different possibilities of hyper-parameters which could influence the performance of embeddings and proposes a set of 3 different intrinsic tasks by covering more than 9 classical and recent combination of baseline datasets.
\end{enumerate}

The paper is structured as follows: Section 2 covers the recent work related to the work presented in the paper and also highlights the novelty of the paper. Section 3 introduces the models considered for experimentation and as a pre-requisite discusses them in details. Section 4 explains the experimentation setup by discussing the training procedure and dataset. It also covers the different evaluation tasks with their respective dataset and procedure in details. Section 5 discusses the results for both of the models in detail and highlights the suitable internal varieties for future researchers and developers. Finally, a conclusion is provided in section 6.

\section{Related Works}
\label{sec:related-works}

\textbf{Word embedding generation:} The idea to represent words in a continuous vector space has been explored for quite some time because the vectorised continuous embedding provides several advantages over a sparse embedding. First, while the dimension size for sparse embedding method like one-hot encoding is equal to the vocabulary size - which could be in millions for large corpus - for a continuous embedding the dimension is just a hyper-parameter and usually, even small size provides sufficiently good results. Second, while sparse embedding only highlights atomic property, a continuous embedding showcase semantic and syntactical properties making the embeddings an ideal candidate for many downstream natural language processing tasks.  

Over the years several methods have been proposed which tries to improve on multiple aspects like capturing more complex semantic and syntactical properties, providing faster training with better accuracy to even working on a very large scale of text corpus like a complete dump of Wikipedia \cite{almeida2019word}.  These methods can be categorised into two classes. First, the count-based -- where the low dimensional latent features are computed by leveraging global word-context co-occurrence counts from a corpus. These are very often represented as word-context matrices, which in the case of early examples like \textit{Latent Semantic Analysis} (LSA) was represented as "term-document" matrix i.e. row and column represent different words and documents in the corpus respectively and cell entries are the exact counts of occurrence of a word in a document. In contrast, future works proposed symmetrical "term-term" matrix, where each row and column represent one word and cell entries are the count of co-occurrence of the pair of words. While methods like \textit{Hyperspace Analogue to Language (HAL)} used the matrix as it is, other methods like \textit{COALS} \cite{rohde2006improved} proposed pre-requisite transformation like entropy or correlation-based normalization on the matrix before final factorization. These additional operations were done to scale the raw co-occurrence counts which could vary in multiple order of magnitude between rare and frequent words. Another method proposed a square root type transformation in the form of\textit{ Hellinger PCA (HPCA)}\cite{collobert2014word} has been suggested as an effective way of learning word representations. Lastly, \textit{GloVe} \cite{pennington2014glove} model proposed that ratios of co-occurrences, rather than raw counts, encode actual semantic information about pairs of words. It also used weighting functions to make sure that not all co-occurrences are equally weighted, especially the ones that occur rarely.

The second class of methods are the shallow window-based -- where a sliding window approach is applied on a large corpus of text and embeddings are learned by considering either the target or context words within a predefined window as input and predicting the other as output. The shallow learning is formulated as a prediction task which considers multiple instances of the local context of words in contrast to count-based methods which uses global count statistics.  Word2vec \cite{mikolov2013efficient} being one of the classical window-based approaches, introduces two different training methods, \textit{Continuous Bag of Words} (CBOW) - where context words are input and target word is predicted as output and \textit{SkipGram} (SG) - where target word is input and context words are predicted as output. The word embeddings are derived as a by-product of the prediction task by extracting the weights of the neural network used to model the \textit{(target word, context word)} prediction problem. This idea was further enhanced by another paper \cite{mikolov2013distributed} by introducing negative sampling which is a  methodology to probabilistically select fewer negative training examples than originally required. This greatly reduced the number of training examples required by the system to converge. To enhance the morphological learning of the word embeddings, FastText \cite{bojanowski2017enriching} proposes learning embeddings for n-gram tokens and using their combination to create word embeddings on the fly. This made it possible to create embedding of words not seen in the training data as well as enhanced performance in NLP tasks for morphologically rich languages. 

For our model analysis task, we have picked GloVe and Word2Vec as the representative of their respective classes. Both of the methods provide high accuracy and are considered as a baseline and first solution for many NLP tasks. FastText was not considered due to two reasons. First, because its significant enhancement to word2vec is only evident in languages with rich morphology. And as we limit the paper to consider only the English language - which is comparatively not morphologically rich - the method could be ignored in favour of word2vec. Second, as FastText learns n-gram representation, this leads to a drastic increase in the memory and computation requirement. With a little gain in accuracy for a larger training time and compute requirement, FastText was dropped from further analysis.

\textbf{Word embedding evaluation:} Quantitative evaluation of word embeddings can be done by performance test on several open-source datasets. The tests can be categorised into two types of evaluation tasks \cite{schnabel2015evaluation}. First, Extrinsic evaluations - where word embeddings are directly used as input features for downstream applications and performance score is the accuracy specific to each task. Another possibility is to train embeddings specific to each task which leads to the creation of a very specific embedding suitable to only one type of downstream application. Some examples could be part-of-speech tagging and named-entity recognition \cite{pennington2014glove}. Second, Intrinsic evaluations - where the training of embedding is unsupervised and not related to the evaluations metrics at all. The trained embeddings are then tested for syntactic or semantic relationships between words by using open-source linguistic datasets. Within the intrinsic evaluation,  different specific evaluation tasks are present such as similarity, association and analogy.  

Even with the availability of several datasets for quantitative evaluation, one major problem area is the selection of different internal varieties of models (model types) for the evaluation. This is especially valid for any dual vector space modelling methods, like word2vec and Glove, where two word weights (word embeddings \textbf{W} and context embeddings \textbf{C}) are trained and the final embeddings are picked by either selecting one of the two \cite{mikolov2013distributed} or taking simple summation \cite{pennington2014glove}. \cite{hill2014not} suggests that not all neural embeddings are created equal. This notion is further discussed by \cite{mitra2016dual} which encourages to look at both the \textbf{W} and \textbf{C} vectors separately - within a single method - as they provide advantages for different tasks. 

While the method and evaluation techniques discussed above provides us with sufficient tools to answers long-standing questions of "\textit{when to use what?}" in word embeddings, they originally themselves fail to cover the wide variety of analysis. Probably the most similar to our work is  \cite{asr2018querying} which demonstrate the usefulness of the dual embeddings for similarity and relatedness tasks. But they only considered a small subset of relatively older datasets and completely overlooked the \textit{CBOW} training method of \textit{word2vec}. Another related recent work is \cite{jatnika2019word2vec} which performs model analysis only on the \textit{CBOW} model. It only considers a small range of context window size and doesn't consider different model types in its evaluations at all.

This paper tries to cover the gaps of the recent word embedding evaluation research by performing an extensive analysis of the performance of word embeddings. This is done by evaluating the unsupervised trained word embeddings on several datasets spread across the three diverse intrinsic evaluation tasks - similarity, association and analogy. We also consider hyper-parameters which are tuneable variables like context window size and embedding dimensions size, with wide range and instances. Different Training methods like CBOW and SG in case of \textit{word2vec} are also considered as they have been shown to provide different task-specific performances. This analysis is further enhanced by considering one representation embedding techniques each from count-based and window-based approach to provide a complete analysis.

\section{Models}
\label{sec:models}
This paper selects one model each from the two different classes - \textit{Glove} from count-based methods and \textit{Word2ec} from window-based prediction methods. We will discuss each technique in details and present a holistic view of the empirical comparison. 

\subsection{Word2Vec}
Word2Vec was introduced in a series of papers \cite{mikolov2013efficient, mikolov2013distributed} as a distributed neural technique to learn word embeddings. While the idea of representing words as continuous value vectors was not new \cite{mnih2009scalable, mikolov2007language, mikolov2009neural}, Word2Vec trained high-quality word embedding with better accuracy on intrinsic word similarity and analogy tasks than existing techniques at the time. It also loosens the computational restraints and made it possible to learn representations for large-sized unstructured corpus consisting of millions of unique vocabulary words. 

\cite{mikolov2013efficient} proposed a very basic 3 layer deep neural network including the input, hidden and output layer. The size of the input and output layer is equal to the size of vocabulary $V$ and the hidden layer size is $D$ which is also the dimension of final word embedding. 
The weights between input-hidden layer can be called word vectors and represented by \textbf{W} similarly, weights between hidden-output layer can be called context vectors and represented by \textbf{C} \cite{asr2018querying}. The idea behind training is to pass one word (in form of 1 hot-encoded vector) as input and predict another word as output (again as 1 hot-encoded vector). The model trains by updating the weights of the layers based on the correct/incorrect predictions. The training data (input-output word pairs) is prepared by sliding a window (of length $w$) across the corpus of large text - which could be articles or novels or even complete Wikipedia data dump. For each such window, the middle word is the target word and the remaining words in the vicinity are context words. With this in mind, two different training methodologies were suggested - (1) \textit{CBOW}:  where the inputs are the context words and the expected output is the target word, and (2) \textit{SG}: which is just the reciprocal of CBOW and takes target word as input and learns to predict the context words. At each iteration of training, we have some positive examples - pairs of target and context words and negative examples - pairs of target and all non-context words. There is a serious computation problem as for a large-sized vocabulary we will have to update all of the non-context words for each training, which will drastically slow down the training procedure. To mitigate this, \cite{mikolov2013distributed} proposed to trim the negative samples by randomly (dependent on word frequency) selecting some negative samples. This proposed technique called "negative sampling" further enhanced the training process, making it possible to train on much larger data without much adverse effect on the embedding quality. The cost function for the SG method with negative sampling is expressed as, 

$$
J_{w2v}=\log \sigma\left(w_{i}^{T}{ } \tilde{w}_{j}\right)+\sum_{k=1}^{m} \mathbb{E}_{w_{k} \sim P_{n}(w)}\left[\log \sigma\left(-{w_{i}^{T}}{ } \tilde{w_{k}}\right)\right]
$$

where $\sigma(.)$ is the sigmoid function. $w_{i}^{T}$ and $\tilde{w}_{j}$ are the word embeddings of the target and context words from the \textbf{W} and \textbf{C} weights respectively. $\tilde w_{k}$ is the embedding of a word outside of the context of the target words $w_{i}$, sampled from a set of all such possible words $P_{n}(w)$. This sampling is of size $m$ and simulates negative sampling. The complete objective functions operate on two aims, first to increase the dot product of the target and context word which is analogous to making the vectors similar. This is represented by the left side of the cost function - where frequently seen pair of target and context word will lead to high dot product value, which the sigmoid function will transform to a value close to 1 and which is finally scaled near 0 by the log function. This makes the cost of an ideal target and context word pair approximate 0. Next, for the pairs of target word and negative samples, the ideal state will be if they have very less similarity hence very small dot product, a negation of which will make it a large value. Then the sigmoid will make it close to 1, which the log will approximate to 0. Hence as expected we get small cost when target and context words are highly associative and the negative samples are not so.

\cite{rong2014word2vec} provides an intuitive understanding of the complete training process where vectors of positive pairs of target and context words from $W$ and $C$ weights are "pulled towards" each other, whereas for non-positive pairs the vectors are "pushed apart". In doing so, words that appear in similar contexts get pushed closer to each other within the $R^d_W$ embedding space (and also within the $R^d_C$embedding space). This leads to an interesting possibility of using any of the two trained weights or even a combination of both for evaluations. For any given evaluation task where for a given cue word $w_{cue}$ we have to find a response word $w_{res}$, we represent WC as taking the vector of $w_{cue}$ from word vector space $R^d_W$ and querying on the context vector space $R^d_C$ to finding the most similar word. Similar analogies can be drawn for other representations such as WW, WC, CW and CC. Based on the intuition from the training procedure, we can hypothesise \cite{mitra2016dual} that the WW (or the CC) representation will give higher similarity for words which have similar context and hence could be synonyms, whereas the WC will be good for word pairs which represent an association or cooccurrence like behaviour. Contrary to all these possibilities, the original paper and several subsequent research suggests picking $W$ as the final word embedding (hence WW comparison method), completely discarding the $C$ embedding.

\textbf{Experimentation:} With intention of testing the performance of different word comparison methods in addition to other tunable hyperparameters, we propose following extensive analysis on the word2vec method. We test different varieties of the model with following variables, (1) Training methods: CBOW vs Skipgram; (2) Word comparison methods: WW, WC, CW, CC; (3) Context window size: 5 and 50, and (4) Word embedding dimensions: 100, 200 and 300. This will leads to a total of 48 different models which we test over the intrinsic evaluation datasets and compare results. 

\subsection{GloVe}
Glove is a global log-bilinear method for word embedding generation \cite{pennington2014glove}, where training utilizes the complete co-occurrence information rather than the windowed context as done in word2vec, hence the name GLObal VEctor representation. It was an extension to the previous global matrix factorization methods like Hellinger PCA \cite{collobert2014word} and PPMI global matrix factorization method \cite{bullinaria2007extracting} with the intent to provide better results for analogy tasks as usually, local context-based approaches like word2vec provide better score in analogy tasks. To capture this behaviour Glove proposes two major modifications, first to use the ratios of co-occurrence probabilities rather than the exact probabilities itself for the word embedding training and second by introducing a weighted function which gives small weights to rare co-occurred word pairs, making sure not all co-occurrences are treated equally. The final cost function formulation for Glove was, 

$$
J_{glove}=\sum_{i, j=1}^{V} f\left(X_{i j}\right)\left(w_{i}^{T} \tilde{w}_{j}+b_{i}+\tilde{b}_{j}-\log X_{i j}\right)^{2}
$$

where $w_{i}^{T}$ and $\tilde{w}_{j}$ are the word embeddings from W and C embeddings, $b_{i}$ and $\tilde{b}_{j}$ are the bias terms and  $ X_{i j}$ is the co-occurrence count of any two words $i$ and $j$ in the vocabulary respectively. The weighted function is further defined as

$$
f(x)=\left\{\begin{array}{cc}
\left(x / x_{\max }\right)^{\alpha} & \text { if } x<x_{\max } \\
1 & \text { otherwise }
\end{array}\right.
$$

with suggested $x_{max}=100$ and $\alpha=3/4$. This creates a clear boundary for word pairs which are seen together more than 100 times as good pairs and the remaining ones as relatively bad ones. Instead of setting a cut-off in training for bad pairs, a non-linear weighting function is applied which decreases as the pair co-occur less and less, finally becoming 0 if they are never seen. Finally, the Glove cost function can be understood as a difference in the dot product of the word embeddings of two words and their log co-occurrence value, which is then weighted by a non-linear function by the rarity of the co-occurrence. The aim is to make the dot product of embeddings of two positive words pairs to be similar to the logarithm of their co-occurrence value. Similar to the word2vec methodology, the embeddings of similar words are "pushed towards" each other. 
Glove also trains two weight embeddings for each word during the training process. This is done to handle overfitting which may arise otherwise. Following the predefined terminology, we refer to them as word vectors $W$ and context vectors $C$. It is suggested \cite{pennington2014glove} that the weights are similar to each other if the $X$ is symmetric. Hence the final values of the W and C weights depend on their random initialization. With this in mind, the original paper suggests taking the \textit{SUM} of both weights as the final word embedding. While this may work in some cases, it is not true for the majority of real-world applications as usually, the relationship of words are directional in nature. For examples, \textit{storks} are more likely to be associated with \textit{baby} than the other way around. To fairly compare the suggestions of the original paper, we also consider the word comparison methods of AA where $A=mean(W, C)$ and SS where $S=sum(W, C)$.

\textbf{Experimentation}: We propose an extensive analysis of the Glove method. Here the variables are, (1) Word comparison method: AA, SS, WW, WC, CW, CC, (2) Context window size: 5 and 50, and (3) Word embedding dimensions: 100, 200 and 300. This leads to a total of 36 different models which we test over the different intrinsic evaluation and compare results.

\section{Experimentation setup}
\label{sec:experimentation Setup}

\subsection{Dataset and process for training}
Word embedding model training requires large corpus of text. It is ideal to keep it as large and diverse as possible to provide sufficient instances of each word in different contexts. To perform comparison across different models, it is also natural to train all models on the same data. To fulfil these requirements, Wikipedia 2017 dump \footnote{https://github.com/RaRe-Technologies/gensim-data/releases/tag/wiki-english-20171001}
was selected which in its entirety contains approximately 5 million diverse articles.  Due to computational reasons, only the first 1 million articles were selected, which contains more than 9 million unique words. The Wikipedia article text in its exact format is not suitable for word embedding training, so following steps of pre-processing were performed. First, articles were further segregated into sections as commonly used in Wikipedia. Several sections like "Further reading", "External links" and "References" were removed as they mostly contain hyperlinks and does not have text in the similar free flow nature when compared to the majority of other sections. Second, the remaining sections were further divided into sentences and subsequently into word tokens. Finally, stopwords tokens were removed, punctuations were deleted and the remaining tokens were lemmatized. The prepared data was then passed to the individual model for training with one fixed parameter of $\text{min count}=10$. This parameter directs to only pick the tokens which are observed at least 10 times in the complete corpus. This led to the removal of rare tokens which can be ignored for this experiment. This reduces the overall vocabulary count, which in turn enhances computation time and further reduce trainable parameters count. 

\subsection{Dataset and process for evaluation}

To analyze the performance of each model we perform intrinsic evaluations. These tasks directly test for syntactic or semantic relationships between words. These tasks typically involve a pre-selected set of query terms and semantically related target words. The performance of a model is measured either by comparing against the provided similarity scores or hit-miss ratios. Another widely used technique is extrinsic evaluations, where as suggested in \cite{schnabel2015evaluation}, the word embeddings are taken as input features to downstream tasks like pos-tagging, entity extraction, sentiment analysis, etc. The performance of the model in this case is measured based on how accurate the downstream application is for a given model's embedding. That said, intrinsic evaluation is considered the baseline in most of the previous model analysis works \cite{gladkova2016intrinsic, schnabel2015evaluation} and also extrinsic evaluation techniques are more expensive when compared to intrinsic evaluations as suggested in \cite{thawani2019swow}. Due to these reasons, we stick with our decision of using Intrinsic evaluations to compare different models. 

We further divide the intrinsic evaluation into three different sub-tasks,  each testing a model on a different type of evaluation domain and respective datasets. In total there are 9 different datasets. The three evaluation sub-tasks are, 

\subsubsection{Similarity evaluation}
Semantic similarity is the comparison of interchangeability of a pair of words. This can be considered as "is-similar-to" kind of relationship between the pairs. Several datasets which are prepared for this kind of evaluation provide a similarity score to determine the confidence of this kind of relationship. One such example is \textbf{WS-353-SIM} \cite{agirre2009study} dataset which scores pairs of words between 0-10 based on how interchangeable they are. Here pairs like \textit{coast} and \textit{shore} has a similarity score of 9.11 whereas \textit{coast} and \textit{hill} only has score of 4.38. Another similar dataset is \textbf{SimLex-999} \cite{hill2015simlex} which in addition considers part of speech and comprises of 666 Noun-Noun pairs, 222 Verb-Verb pairs and 111 Adjective-Adjective pairs. \textbf{RG-65} \cite{rubenstein1965contextual} contains 65 noun pairs with similarity scores. For variety we also consider \textbf{SimVerb-3500} \cite{gerz2016simverb} and \textbf{EN-VERB-143} \cite{baker-etal-2014-unsupervised}, which contains 3,500 and 143 pairs of verbs, respectively, with similarity score bounded between 0 and 4. This takes the total count of datasets for similarity evaluation to 5 and their coverage is diverged in terms of part of speech, scoring metric range and collection timeline. 

The datasets also contains a mix of \textit{True Positives (TP)} and \textit{True Negative (TN)} examples. For the majority of the word pairs with high similarity score i.e. TP, we also have examples for the same cue word such that it is a part of another pair with fairly low similarity score i.e. TN. This diverse collection makes it possible to use a correlation metric for the evaluation of a model. In this case, the cosine similarity of the embeddings of the word pairs can be considered as the model's confidence of similarity between the word pairs. Correlating this with the actual score present in the dataset and doing this for all positive and negative examples will highlight the model's performance for the similarity task. This will showcase if the learned embedding spaces capture some sense of similarity or not. When comparing multiple models against each other, a higher correlation score means it has captured the sense of similarity better. For this pearson correlation can be used and the semantic evaluation score of a model $m$ for a dataset $d$ with $k$ word pairs can be computed by, 

$$
\operatorname{corr({m},{d})}=\frac{n\left(\sum x_{k} y_{k}\right)-\left(\sum x_{k}\right)\left(\sum y_{k}\right)}{\sqrt{\left[n \sum x_{k}^{2}-\left(\sum x_{k}\right)^{2}\right]\left[n \sum y_{k}^{2}-\left(\sum y_{k}\right)^{2}\right]}};
\\
$$

$$
\text{where,   } \space \space
\operatorname{x_{k}}=cos(w_{i}^m, w_{j}^m)=\frac{w_{i}^m \cdot w_{j}^m}{||w_{i}^m||\space ||w_{j}^m||}
$$

here $x_k$ is the cosine similarity and $y_k$ is the dataset similarity score for the $k^{th}$ word pair $w_i$ and $w_j$.

\subsubsection{Association evaluation} 
Semantic association is different from semantic similarity as while the later is analogous to interchangeability, the former is analogous to relatedness. For this type of analysis cue-response type of datasets were picked, where for a given cue word, human annotators were asked to provide response words with high relatedness relationship. One such dataset is \textit{Small world of words (SWOW)} \cite{de2019small} which is a collection of more than $12,000$ unique cues and totalling 1 million cue-response entries. For a given cue, each annotator was asked to provide 3 responses one after another namely R1, R2 and R3. In this dataset, for each cue, we compute an R123 score by considering all of the responses among the 3 responses categories. This score denotes the ratio of occurrence of a response when compared against all other responses for the given cue. Later R123 score was used to prune responses with score less than 10 \%. This brought down the unique cues to $\approx8500$ and responses to $\approx12,000$  and led to the formation of \textbf{SWOW-8500} \cite{thawani2019swow}.  Another dataset is \textbf{EAT} \cite{kiss1973associative} which contains aggregation of responses with their strength score for a total of $\approx8200$ cues. By pruning the noisy responses, i.e. with a strength less than 10 \% , we are left with $\approx6500$ cues and $\approx13,000$ responses. 

The processed datasets contain a list of potential responses for any given cue. Association evaluation is then formulated as a combination of hit ratio and coverage of the human-annotated responses $H$ and model computed responses $M$, for a cue. Model computed responses were generated by taking cosine similarity of a cue with every word in the model vocabulary and finding top $N$ most similar words. After several trial and error, the value of $N$ was kept 10. While hit ratio signifies the number of cues for which there is at least 1 overlap between $H$ and $M$ i.e. $H \cap M \neq \emptyset $, coverage is the mean overlap of the $H$ and $M$ across all cues.

\subsubsection{Analogy evaluation} 
Analogy evaluation method, also called Analogical reasoning, is based on the idea that human's notion of cognitive reasoning of relationship can be captured by word embeddings. Given a set of three words, $a$, $a*$ and $b$, the task is to identify another word $b*$ such that the relation $b$:$b*$ is the same as the relation $a$:$a*$. For instance, if one has words $a=London$, $a*=England$, $b=New Delhi$, then the target word would be $India$ since the relation $a:a*$ is $capital:country$. Each such combination of 4 tokens can be termed as an analogy question where the intention is to take $a$, $a*$ and $b$ and predict $b*$. Dataset \textbf{BATS} (acronym for \textit{Bigger Analogy Test Set}) \cite{gladkova2016analogy}, contains thousands of partial analogy questions (only first 2 terms i.e. $a$ and $a*$) divided into 4 classes (\textit{inflectional morphology}, \textit{derivational morphology}, \textit{lexicographic semantics} and \textit{encyclopedic semantics}) and 10 smaller subclasses. For our experiment, questions from \textit{inflectional morphology} section were ignored as models were trained on lemmatized tokens and hence could have high miss rate for the questions from this section. For the remaining sections, partial questions from the same subclass were joined randomly to create a complete analogy question. This led to a total of  $440$ complete questions. Another dataset \textbf{Google Analogy}  \cite{mikolov2013efficient} contains $19,544$ questions divided into 2 classes \textit{morphological relations} with $10,675$ questions and \textit{semantic relations} with $8,869$ questions. On random sampling $\approx1000$ questions were selected for evaluations.

Contrary to similarity and association evaluations, where a query was a single cue, here we have three word tokens in the query i.e. $a$, $a*$ and $b$. \cite{levy-goldberg-2014-linguistic, levy2015improving} suggest to use 3COSMUL metric to find the answer token $b*$ as it mitigates the "soft-or" behaviour of original 3COSADD metric used in the \cite{mikolov2013distributed} for analogy related task evaluating. This generates a score for every word in the vocabulary which denotes its possibility of being the answer token. We pick the top $N=3$ tokens as the potential list of answers and a question is marked as answered if this potential list contains the correct answer. The overall performance score is the average of the computed hit ratio. The 3COSMUL score is computed by,

$$
 \operatorname{3COSMUL(a, a*, b, b*)}=\frac{\cos \left(b^{*}, a^{*}\right) \cdot \cos \left(b^{*}, b\right)}{\cos \left(b^{*}, a\right)+\varepsilon}
$$

where $cos(\cdot)$ is the cosine similarity function and $\varepsilon = 0.001$ is used to prevent division by zero error.

\section{Results}
\label{sec:results}

All different models were trained and later evaluated on the datasets as suggested in the last section. As the number of models and their report on evaluation types and subsequent datasets could be vast and difficult to comprehend, a consolidated statistics is required. 
For this first, the performance scores for each dataset were averaged to provide an aggregated final score for each evaluation type. Then, maximum and average scores were generated by aggregating over the word embedding dimension variable and keeping the rest of the variables intact as they were. 

\subsection{Word2Vec report} 
The consolidated report for word2vec covering all three evaluation types w.r.t. the variables - training method, context window size and comparison method is provided in Table 1. As suggested for word2vec \cite{mikolov2013distributed}, the default compare method is WW, training method is skipgram and context window is 5, hence we represent this model as \textit{(WW, sg, 5)} and consider it as the baseline which is to be compared against other word2vec model variations.
For similarity evaluation, the baseline provides the best maximum score but the best average score is reported by \textit{(WW, cbow, 50)} model. Apart from WW model type, \textit{(WC, cbow, 50)} and \textit{(CW, cbow, 50)} also shows comparable high scores. This hints towards the preference of cbow with high context window for semantic similarity-based tasks. One interesting observation is the "flipping" behaviour when comparing cbow and sg methods. For the former, large context window of 50 is the majority best performer across compare method whereas for the latter the behaviour is flipped and a small context window of 5 is the majority best performer. This suggests the use of small and large context window with sg and cbow respectively for the similarity evaluation tasks. This flipping behaviour can be observed in the visualization of the result in Figure 1, where we haven't aggregated across the embedding dimension variable.
For association task, \textit{(WC, cbow, 50)} model overthrows the baseline by quite a margin. Similar good scores in the same compare method suggest the preference of WC compare method for association or relatedness evaluation tasks. This behaviour is expected as during the training procedure of the word2vec, target and context word embeddings are pushed towards each other but in their respective embedding space. So considering both the embedding spaces when looking for related words similarity seems to be the obvious choice. It is also to be noted the unusually bad results when using CW compare method. One reason for this behaviour as suggested by \cite{asr2018querying} is due to the fact that relatedness between words is directional in nature -  \textit{storks} are more likely to be associated with \textit{baby} than the other way around. The flipping behaviour is observed here as well. Another observation is regarding the lower dependency on the context window size, as contrary to the similarity evaluation task's result. Change in context window size doesn't lead to a drastic change in evaluation score.
For analogy task, \textit{(WW, cbow, 5)} model outperforms others. Comparable high scores suggest the preference of WW for the analogy related tasks. One interesting observation is the preference of small context window when looking across compare methods, as it seems to give the majority best results.

\begin{table}[ht]
\centering
\begin{tabular}{ || c | c | c || c | c | c | c ||} 
 \multicolumn{3}{c|}{} & \multicolumn{4}{||c||}{Compare method} \\
\toprule
                    &    &  &        WW &       WC &       CW &      CC \\
Task & Train method & Context window &              &              &              &              \\
\midrule
Similarity & CBOW & 5 &  \underline{0.520 (0.504)} &  0.430 (0.430) &  0.440 (0.423) &  0.419 (0.408) \\
                    &    & 50 &  0.526 (\textbf{0.514}) &  \underline{0.516 (0.505)} &  \underline{0.510 (0.506)} &  0.348 (0.343) \\
                    & SG & 5 &  \textbf{0.528} (0.511) &  0.476 (0.466) &  0.471 (0.466) &  0.499 (0.478) \\
                    &    & 50 &  0.424 (0.418) &  0.420 (0.398) &  0.410 (0.395) &  0.387 (0.373) \\
                    \hline
                    \hline
Association & CBOW & 5 &  0.148 (0.147) &  \underline{0.244 (0.236)} &  0.001 (0.000) &  0.080 (0.066) \\
                    &    & 50 &  0.191 (0.176) &  \textbf{0.274 (0.254)} &  0.000 (0.000) &  0.081 (0.074) \\
                    & SG & 5 &  0.127 (0.122) &  \underline{0.242 (0.237)} &  0.089 (0.081) &  0.113 (0.107) \\
                    &    & 50 &  0.123 (0.113) &  0.162 (0.153) &  0.040 (0.033) &  0.121 (0.104) \\
                    \hline
                    \hline
Analogy & CBOW & 5 &  \textbf{0.529 (0.494)} &  0.364 (0.351) &  0.012 (0.004) &  0.410 (0.404) \\
                    &    & 50 &  \underline{0.495 (0.462)} &  0.454 (0.433) &  0.000 (0.000) &  0.411 (0.404) \\
                    & SG & 5 &  \underline{0.513 (0.476)} &  0.418 (0.382) &  0.350 (0.309) &  \underline{0.509 (0.460)} \\
                    &    & 50 &  0.414 (0.373) &  0.339 (0.299) &  0.265 (0.213) &  0.400 (0.369) \\
\bottomrule
\end{tabular}
\caption{Word2Vec consolidated report for the three evaluation tasks. The result scores are in the format of "maximum (average)" of the scores when aggregating on the embedding dimensions of 100, 200 and 300. Bold marks the best result and underline highlight comparable high results.}
\end{table}

\begin{figure}[h]
\includegraphics[width=\textwidth]{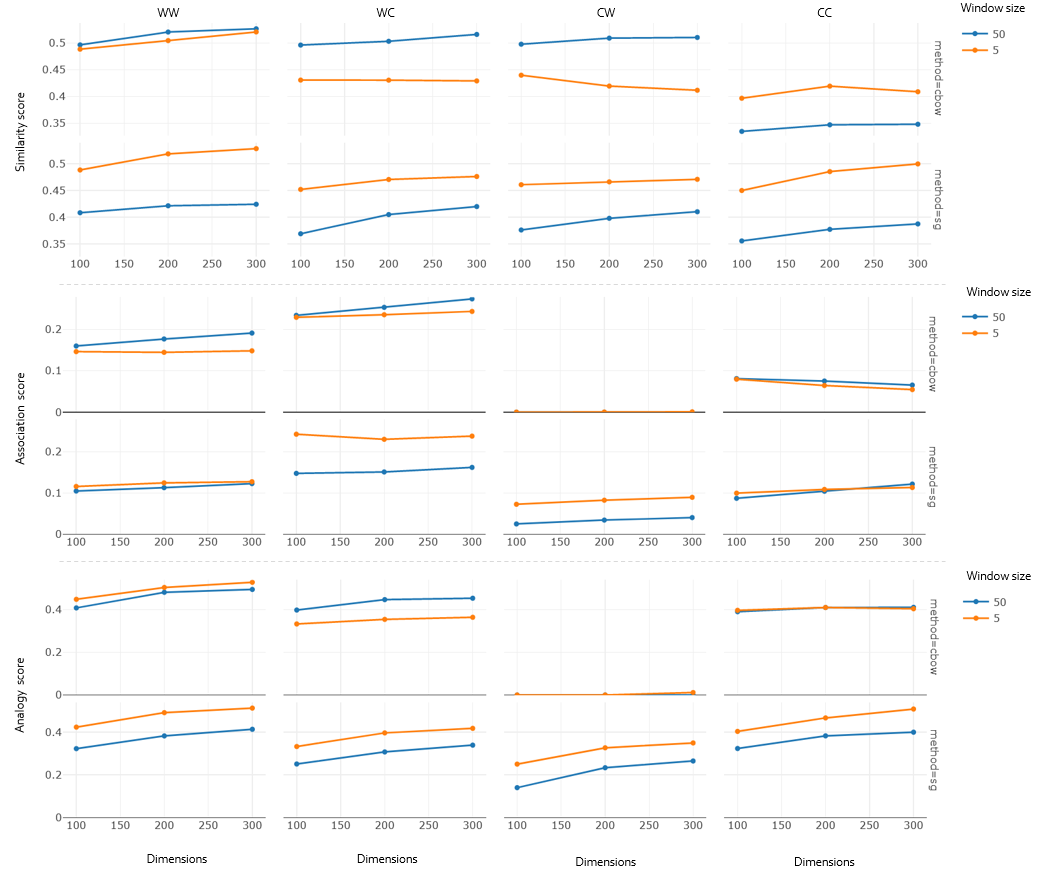}
\caption{Word2vec report visualization for the three evaluation tasks - highlighting the variation in score across embedding dimensions, training method and context window size.}
\centering
\end{figure}

\subsection{Glove report} 
The consolidated report for Glove is provided in Table 2 along with the visualization in Figure 2. As the main paper \cite{pennington2014glove} suggest to use SS compare method and small context window, we will consider \textit{(SS, 5)} as the baseline which is to be compared against other Glove model variations. Note, we have also performed an evaluation on AA compare method, but dropped it from the result table for the sake of brevity as the scores were almost similar to SS. Contrary to the expectation, it is interesting to note that WW and CC  outperform the baseline SS in all of the tasks. In case of similarity evaluations, \textit{(WW, 5)} is the clear winner with \textit{(CC, 5)} and \textit{(SS, 5)} coming 2nd and 3rd respectively. For association evaluation, a similar behaviour is observed. Here while the average score is a tie between  \textit{(WW, 5)} and  \textit{(CC, 5)}, the former performs marginally better in terms of the maximum score. Lastly, for the analogy evaluation, we see a change in preference to larger window size.  \textit{(CC, 50)} scores highest with \textit{(WW, 50)} coming 2nd. 
In case of Glove, we observe WC and CW as the anomalies for every evaluation task, as they are the worst performer across the tasks. This suggests the lower dependence between the embedding spaces and either one or an aggregation of the embedding vectors can be used to get better results. 
Also, we observe a compare method specific flipping behaviour especially in the case of similarity and association evaluation. For WC and CW, higher performance is observed for larger window length. This is in complete contrast to the rest of the compare methods where smaller window length has better performance. 
Overall CW model is the worst performer and WW model is the best one with the highest score in 2 out of 3 evaluation tasks and a close 2nd in the last one.  
Next considering only the high performing models i.e. SS, WW and CC, two interesting observations can be made when comparing similarity and association evaluations again the analogy evaluations. In the case of former, there is a significant difference in the performance of models w.r.t. context window size and smaller window size usually dominates. While in case of later, the behaviour is flipped and both the context window sizes provide somewhat similar results with larger window size providing slightly better scores across all compare methods.

One point to note is the difference in performance when comparing word2vec and Glove at a whole, with word2vec performing better. As suggested by \cite{pennington2014glove}, it's due to the difference in the training procedure of both methods, making it difficult to simulate similar training and hence perform a bipartisan evaluation. This paper focused more on the behaviour of internal model variations for each methods and hence a holistic across method comparison is out of the scope. That said, all models were trained with similar training time in mind, and hence it showcases word2vec's behaviour of training faster for better results when compared to Glove. 

\begin{table}[ht]
\centering
\begin{tabular}{|| c | c || c | c | c | c | c ||}
\multicolumn{2}{c|}{} & \multicolumn{5}{||c||}{Compare method} \\
\toprule
                    &  &            SS &               WW &         WC &         CW &        CC \\
Task & Context window &                &               &                &                &                \\
\midrule
Similarity & 5 &  \underline{0.479 (0.464)} &   \textbf{0.494 (0.473)} &  0.394 (0.388) &  0.401 (0.395) &  \underline{0.484 (0.467)} \\
                    & 50 &  0.460 (0.443) &  0.455 (0.435) &  0.433 (0.420) &  0.430 (0.423) &  0.457 (0.434) \\
                    \hline
                    \hline
Association & 5 &  0.192 (0.189) &  \textbf{0.204 (0.196)} &  0.131 (0.112) &  0.130 (0.111) &  \underline{0.202} (\textbf{0.196)} \\
                    & 50 &  0.193 (0.183) &  0.184 (0.171) &  0.158 (0.145) &  0.154 (0.143) &  0.186 (0.171) \\
                    \hline
                    \hline
Analogy & 5 &  0.459 (0.434)  &  0.478 (0.446) &  0.222 (0.211) &  0.216 (0.205) &  0.484 (0.457) \\
                    & 50 &  0.489 (0.460) &  \underline{0.491 (0.462)} &  0.342 (0.327) &  0.329 (0.321) &  \textbf{0.496 (0.462)} \\
\bottomrule
\end{tabular}
\caption{Glove consolidated report for the three evaluation tasks. The scores are in the format of "maximum (average)" when aggregating on the dimensions of 100, 200 and 300. Bold marks the best result and underline highlight comparable high results.}
\end{table}

\begin{figure}[h]
\includegraphics[width=\textwidth]{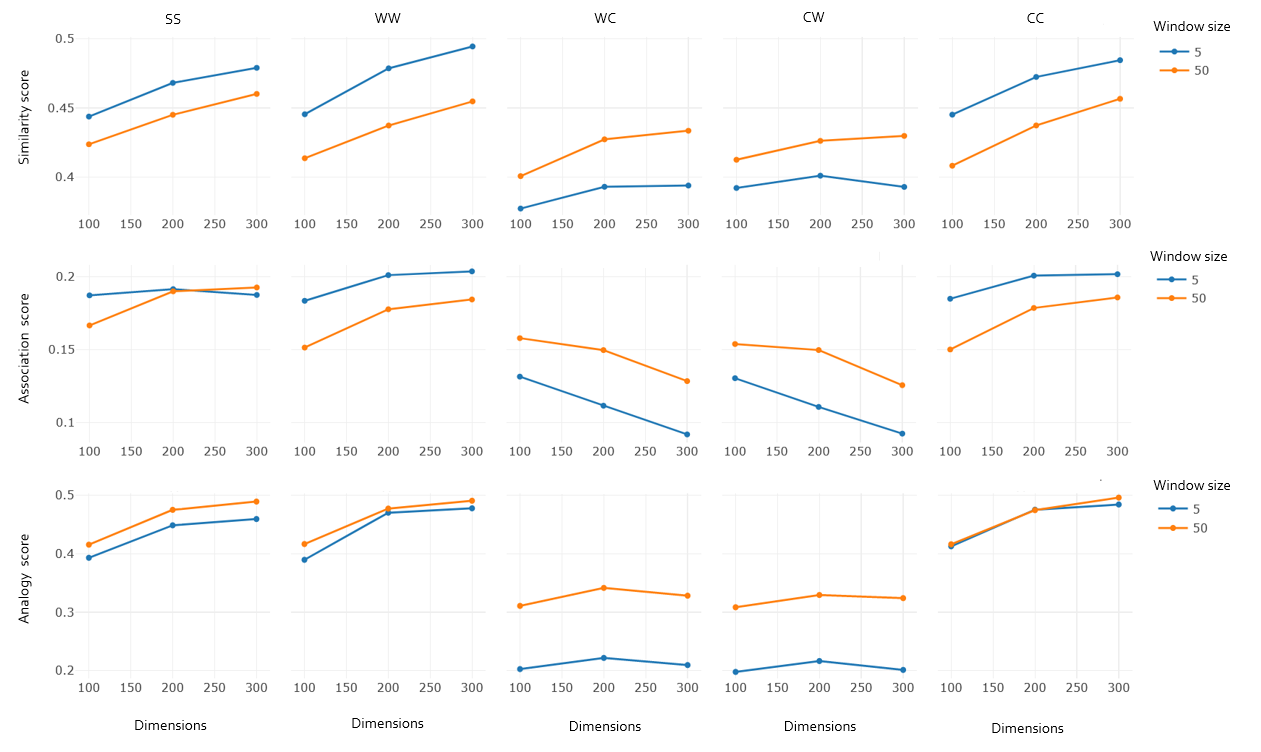}
\caption{Glove report visualization for the three evaluation tasks - highlighting the variation in score across embedding dimensions and context window size.}
\centering
\end{figure}

\section{Conclusion}
Recently the majority of the research has been focused on proposing sophisticated newer models to train word embeddings. While newer models do provide better accuracies, they use classical models with default hyper-parameters as a baseline to showcase enhancement. On the other hand, there has been little work on evaluating the different relevant varieties of the existing models. While some variety is covered by hyper-parameters such as embedding dimension, context window size and training method, another interesting class of variety emerges when we consider methods with dual embedding spaces. This leads to a variable called compare method, where we combine or perform directed comparison between two embedding weights to replicate the cue-response type of tasks. In this paper, we consider all of these variations to compare two classical embedding methods belonging to two different methodologies - Word2Vec from window-based and Glove from count-based. The comparative evaluation was done by scoring different models across three different evaluation tasks covered by 9 open-source linguistic datasets. The finding is present in the form of consolidated reports table and visualization charts.  The overall report showcase the preference of non-default variables in the majority of the tasks for both of the embedding methods, which is in line with \cite{asr2018querying}.  The generated reports are proposed as a lookup reference sheet which highlights the choice of model varieties which perform better for different intrinsic tasks. This can further be connected to similar extrinsic evaluations.  This can greatly enhance the fundamental research baselining and applied downstream application performance. 

\printbibliography


\end{document}